\documentclass[letterpaper, 10 pt, journal, twoside]{ieeetran}
\usepackage[utf8]{inputenc}
\DeclareUnicodeCharacter{FF0C}{,}
\usepackage{amssymb}
\usepackage{amsmath}
\usepackage{amsfonts}
\usepackage{graphicx}
\usepackage{booktabs}
\usepackage{algorithm}
\usepackage{algorithmic}
\usepackage{url}
\usepackage{xcolor}
\usepackage{listings}
\usepackage{float}
\usepackage{caption}
\usepackage{subcaption}
\usepackage{framed}
\usepackage{array}
\usepackage{placeins}
\usepackage{orcidlink}

\usepackage{hyperref}

\hypersetup{
    colorlinks=true,
    linkcolor=blue,
    citecolor=blue, 
    filecolor=magenta,
    urlcolor=blue,
    pdftitle={MTIL: Encoding Full History with Mamba for Temporal Imitation Learning},
    bookmarksnumbered=true,
    bookmarksopen=true,
    bookmarksopenlevel=1,
    pdfauthor={}
}
\begin{document}
\title{MTIL: Encoding Full History with Mamba for Temporal Imitation Learning}

\author{Yulin Zhou\orcidlink{0009-0008-8302-2606}, Yuankai Lin\orcidlink{0000-0001-5764-514X},  Fanzhe Peng\orcidlink{0009-0009-3400-1152},  Jiahui Chen\orcidlink{0009-0002-2471-7180},  Kaiji Huang\orcidlink{0000-0002-0090-3088},  Hua Yang\orcidlink{0000-0002-5430-5630},~\IEEEmembership{Member,~IEEE} and Zhouping Yin\orcidlink{0009-0000-7091-2265},~\IEEEmembership{Member,~IEEE}%
\thanks{Manuscript received: May 18, 2025; Revised: August 9, 2025; Accepted: September 20, 2025.}%
\thanks{This paper was recommended for publication by Editor Aleksandra Faust upon evaluation of the Associate Editor and Reviewers' comments. This work was supported by the Joint Funds of the National Natural Science Foundation of China (Grant No. U22A20208), the Natural Science Foundation Innovation Group Project of Hubei Province (Grant No. 2022CFA018), and the Key Research and Development Program of Guangdong Province (Grant No. 2022B0202010001-2).(Corresponding author: Hua Yang.)}%
\thanks{Y. Zhou, Y. Lin, F. Peng, J. Chen, K. Huang, H. Yang, and Z. Yin are with the School of Mechanical Science and Engineering, Huazhong University of Science and Technology, Wuhan 430074, China (e-mail: yulinzhou@hust.edu.cn; huayang@hust.edu.cn).}%
\thanks{Digital Object Identifier (DOI): 10.1109/LRA.2025.3615520.}%
}

\markboth{IEEE ROBOTICS AND AUTOMATION LETTERS. PREPRINT VERSION. ACCEPTED SEPTEMBER, 2025}%
{Zhou \MakeLowercase{\textit{et al.}}: MTIL: Encoding Full History with Mamba for Temporal Imitation Learning}

\maketitle
\begin{abstract}
Standard imitation learning (IL) methods have achieved considerable success in robotics, yet often rely on the Markov assumption, which falters in long-horizon tasks where history is crucial for resolving perceptual ambiguity. This limitation stems not only from a conceptual gap but also from a fundamental computational barrier: prevailing architectures like Transformers are often constrained by quadratic complexity, rendering the processing of long, high-dimensional observation sequences infeasible. To overcome this dual challenge, we introduce Mamba Temporal Imitation Learning (MTIL). Our approach represents a new paradigm for robotic learning, which we frame as a practical synthesis of World Model and Dynamical System concepts. By leveraging the linear-time recurrent dynamics of State Space Models (SSMs), MTIL learns an implicit, action-oriented world model that efficiently encodes the \textit{entire} trajectory history into a compressed, evolving state. This allows the policy to be conditioned on a comprehensive temporal context, transcending the confines of Markovian approaches. Through extensive experiments on simulated benchmarks (ACT, Robomimic, LIBERO) and on challenging real-world tasks, MTIL demonstrates superior performance against SOTA methods like ACT and Diffusion Policy, particularly in resolving long-term temporal ambiguities. Our findings not only affirm the necessity of full temporal context but also validate MTIL as a powerful and a computationally feasible approach for learning long-horizon, non-Markovian behaviors from high-dimensional observations. Project code are available at \url{https://github.com/yulinzhouZYL/MTIL}
\end{abstract}


\section{Introduction}
\IEEEPARstart{I}{mitation} Learning (IL) has emerged as a powerful paradigm for teaching robots complex skills directly from expert demonstrations, bypassing the need for intricate reward engineering often required in reinforcement learning~\cite{osa2018algorithmic, argall2009survey, schaal1997learning}. Behavioral Cloning (BC), the simplest form of IL, learns a direct mapping from observations to actions via supervised learning and has enabled robots to perform a variety of tasks~\cite{argall2009survey, schaal1997learning, pomerleau1989alvinn, osa2018algorithmic}. Recent advancements, particularly leveraging powerful sequence models and generative approaches, have led to state-of-the-art (SOTA) methods such as the Action Chunking Transformer (ACT)~\cite{zhao2023act, zhao2024aloha} and Diffusion Policy~\cite{chi2023diffusionpolicy, ho2020ddpm, ze2023dp3, pearceimitating}, which excel at learning visuomotor control policies for complex manipulation.
\IEEEpubidadjcol
Despite these successes, a fundamental limitation persists in many current IL approaches: the reliance on the Markov assumption. These methods typically predict the action $a_t$ based solely on the current observation $o_t$ or a short, fixed-length history window $o_{t-k:t}$ ~\cite{osa2018algorithmic, ross2011dagger, mandlekar2021robomimic, chen2021decision}. This assumption breaks down in tasks where the history beyond this limited window is necessary to resolve ambiguity in the current observation. Consider a sequential task requiring a robot to first place an object at location A, and subsequently move it to location B. At an intermediate configuration, the robot's visual observation and proprioceptive state ($o_t$) might be identical regardless of whether it has successfully completed the sub-task at location A. A Markovian policy, lacking the memory of visiting A, cannot distinguish these fundamentally different historical contexts and may erroneously proceed directly to B, failing the task~\cite{osa2018algorithmic, lynch2020learning, gao2024prime}. This temporal ambiguity signifies an underlying Partially Observable Markov Decision Process (POMDP), posing a critical challenge for standard IL methods in state-dependent tasks.

Since ambiguous tasks manifest as POMDPs and human demonstrations are inherently non-Markovian, effective imitation necessitates history-aware policies. While the critical role of historical context has been increasingly recognized~\cite{mandlekar2021robomimic}, a fundamental barrier has remained: the computational infeasibility of processing long, high-dimensional observation histories with prevailing architectures like Transformers. Addressing this, we introduce Mamba Temporal Imitation Learning (MTIL). Our approach is specifically designed to incorporate the complete observational history into decision-making by leveraging the unique properties of State Space Models (SSMs), particularly the recently developed Mamba architecture~\cite{gu2023mamba, jia2024mail}. Mamba's recurrent structure allows it to maintain a compressed hidden state $h_t$ that theoretically encapsulates information from the entire preceding observation sequence $H_t = (o_1,..., o_t)$. Instead of relying solely on $o_t$, MTIL learns a policy $\pi(a_t | h_t, o_t)$ that explicitly conditions the action prediction on this history-infused hidden state $h_t$ in conjunction with the current observation $o_t$, enabling differentiation between observationally similar states and the correct execution of complex sequential tasks.
Our contributions are threefold:
\begin{enumerate}
    \item We propose MTIL, a novel architecture that is the first to leverage the linear-time recurrence of State Space Models to make full-trajectory imitation learning from high-dimensional visual data computationally feasible on commodity hardware, breaking the quadratic bottleneck of attention-based models.
    \item We provide a new theoretical framing for this approach, positing MTIL as learning an \textit{implicit dynamical system}. This system's evolving state acts as a continuous belief-state representation, offering a robust solution to the core problem of temporal ambiguity in partially observable environments.
    \item We provide extensive empirical validation demonstrating that MTIL significantly outperforms state-of-the-art methods, including ACT and Diffusion Policy. Furthermore, on tasks requiring long-term memory, MTIL also surpasses other full-history-capable baselines like Transformer-XL, validating the unique advantages of its underlying architecture.
\end{enumerate}

\section{Related Works}
\label{sec:related_works}
\subsection{Markovian and Short-History Imitation Learning}
A cornerstone of imitation learning, Behavioral Cloning (BC), typically learns a Markovian policy $\pi(a_t | o_t)$ via supervised learning~\cite{argall2009survey, schaal1997learning, pomerleau1989alvinn, osa2018algorithmic}. Although fundamental, this approach inherently struggles with covariate shift and tasks that require memory beyond current observation~\cite{schaal1997learning, ross2011dagger}. Many contemporary methods, despite advances, effectively operate within similar constraints or rely on limited observation histories. For instance, the Action Chunking Transformer (ACT)~\cite{zhao2023act, zhao2024aloha}, leveraging the Transformer architecture~\cite{vaswani2017attention}, predicts chunks of actions $a_{t:t+K-1}$ conditioned on present observation and potentially a latent variable from a CVAE. Although action chunking improves temporal smoothness and reduces the effective horizon~\cite{shi2023awe, fu2024chunking}, its temporal modeling is largely confined to short-term dependencies implicitly captured through the time aggregation of chunks while reasoning, potentially failing when resolving ambiguities requires longer context~\cite{zhao2023act}. Similarly, Diffusion Policy~\cite{chi2023diffusionpolicy, ho2020ddpm, ze2023dp3, pearceimitating}, while adept at capturing complex, multimodal action distributions~\cite{chi2023diffusionpolicy, li2024diffcontrol}, commonly conditions the diffusion process on the present observation or a short history, limiting its capacity for tasks requiring long-term memory~\cite{chi2023diffusionpolicy, li2024diffcontrol}. While extensions like Diff-Control~\cite{li2024diffcontrol} introduce forms of statefulness, they differ fundamentally from MTIL's direct use of a recurrent SSM state to encode the full task history. Other techniques, including Implicit BC~\cite{florence2022implicit} and Energy-Based Models~\cite{bashiri2021distributionally}, also often operate primarily on the current state.

\subsection{Temporal and Sequential Imitation Learning}
The inadequacy of the strict Markov assumption has long motivated efforts to incorporate temporal context. Early explorations employed Recurrent Neural Networks (RNNs) like LSTMs~\cite{osa2018algorithmic, pomerleau1989alvinn, mandlekar2021robomimic, beck2024xlstm}. However, these architectures face challenges with long-term dependencies due to vanishing gradients~\cite{beck2024xlstm}. Furthermore, practical implementations often resorted to fixed history windows and periodic state resets (e.g., sequence lengths of 10-50 in Robomimic~\cite{mandlekar2021robomimic, mandlekar2023mimicgen}), precluding the capture of full trajectory history. More recently, Transformer-based models (e.g., BeT~\cite{shafiullah2022bet}, RT-1~\cite{brohan2022rt1, Haldarbaku}, OPTIMUS~\cite{2023optimus}, ICRT~\cite{2024context}, Baku~\cite{brohan2022rt1, Haldarbaku}, MDT~\cite{2023mdt}) have become prominent, utilizing self-attention to model sequence correlations. {However, the $O(L^2)$ computational complexity of attention imposes practical limits on the size of the context window~\cite{osa2018algorithmic, beck2024xlstm}, hindering their ability to efficiently process entire long trajectories. Even recurrent variants like Transformer-XL~\cite{tianci2024transformerxl}, while theoretically capable of processing long sequences, still rely on the computationally intensive attention mechanism. Distinct strategies for managing long horizons involve temporal abstraction. Hierarchical Imitation Learning~\cite{lynch2020learning, gao2024prime, zhu2022bottom, gupta2019relay, mao2024robomatrix} and Skill Chaining~\cite{lynch2020learning, 2021adversarial, chen2024scar} decompose tasks, learning policies over skills or sub-goals. Waypoint-based methods like AWE~\cite{shi2023awe} or primitive-based approaches like PRIME~\cite{gao2024prime} operate at higher levels of abstraction. While effective, these approaches fundamentally differ from MTIL, which aims to directly model the complete low-level observation-action sequence history, potentially offering robustness against issues like error propagation in skill chaining~\cite{lynch2020learning, chen2024scar}.}

\subsection{State Space Models (SSMs) and Mamba in Robotics}
State Space Models (SSMs) represent a compelling paradigm for sequence modeling, defined by their recurrent hidden state dynamics~\cite{gu2023mamba, jia2024mail, bevandakoopman, fernando2025existence,2025mambaflow, du2025mambaflow, zeng2024mambamos}. Mamba~\cite{gu2023mamba} marked a significant advancement, introducing input-dependent parameters ($\mathbf{A}, \mathbf{B}, \mathbf{C}, \Delta$) via a selective scan mechanism. This allows Mamba models to dynamically focus on relevant sequence information while maintaining the linear time complexity characteristic of SSMs, synergizes the capacity for long-range dependency modeling, akin to Transformers, with the efficient recurrent updates reminiscent of RNNs, yet sidesteps the quadratic scaling bottlenecks of the former~\cite{osa2018algorithmic, beck2024xlstm} and the gradient propagation issues of the latter~\cite{beck2024xlstm}. achieving strong empirical results~\cite{gu2023mamba}.The robotics community has begun investigating Mamba's potential~\cite{gu2023mamba, jia2024mail, tsuji2025mamba, jia2025xil}. For instance, MaIL~\cite{jia2024mail} employed Mamba as an imitation learning backbone, showing promise particularly in low-data regimes~\cite{gu2023mamba}. Mamba Policy~\cite{cao2024mambapolicy, reuss2024edt} integrated Mamba structures within diffusion models to enhance efficiency, while X-IL~\cite{jia2025xil} explored Mamba within a modular IL framework. While these works adeptly leverage Mamba's sequence processing power, MTIL distinguishes itself through its core premise: harnessing the step-updated recurrent state $h_t$ as an explicit, dynamically built representation of the \textit{entire} observation history.This approach, tightly coupled with its sequential training methodology, directly overcomes the temporal ambiguity challenges inherent in Markovian assumptions common in imitation learning.

\section{Mamba Temporal Imitation Learning (MTIL)}
\label{sec:method}
We introduce Mamba Temporal Imitation Learning (MTIL), a novel imitation learning framework designed to overcome the limitations of the Markov assumption by leveraging the full history of observations encoded within the recurrent state of an advanced State Space Model (SSM) architecture.

\begin{figure*}[!htbp]
    \centering
    \vspace{2ex}
    \includegraphics[width=0.9\textwidth]{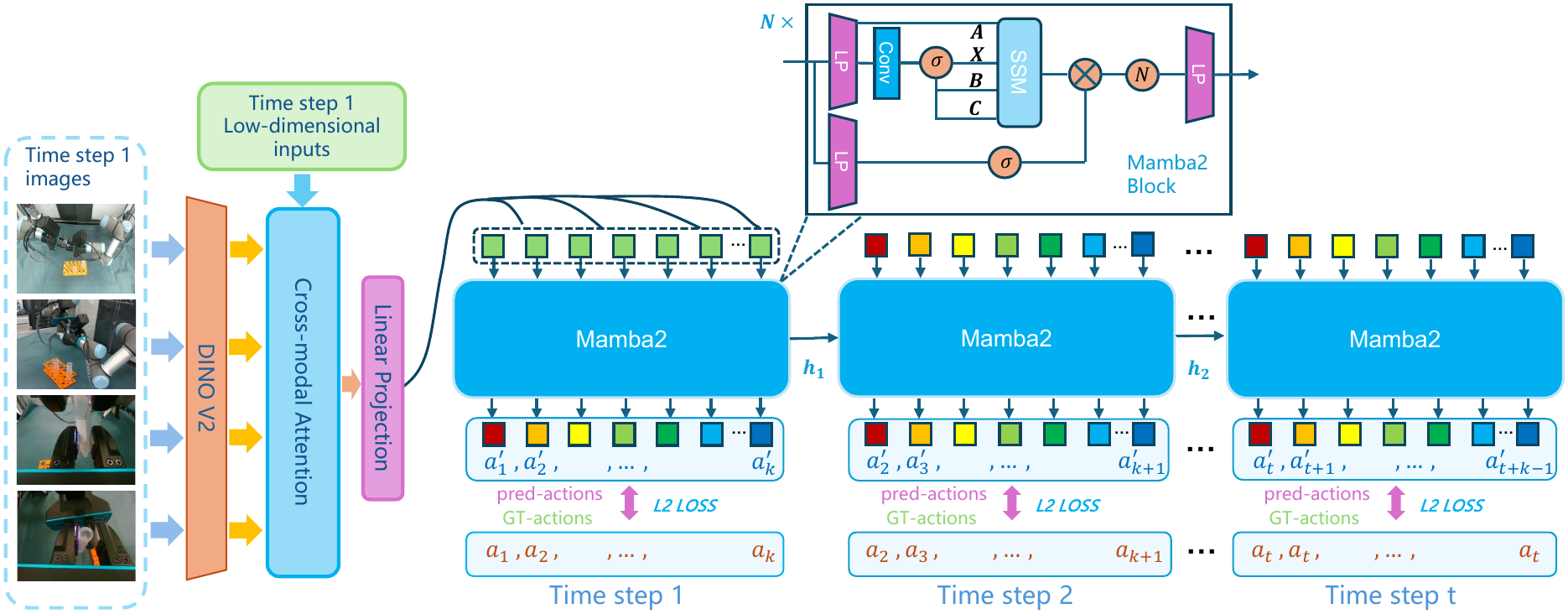}
    \caption{Overview of the Mamba Temporal Imitation Learning (MTIL) architecture. Multi-modal inputs (images via DINOv2, state) are fused and processed by sequential Mamba-2 blocks, updating the recurrent state $h_t$ which encodes history. At each step $t$ across the entire trajectory, MTIL predicts an action chunk $\hat{a}_{t:t+K-1}$ (current plus $K-1$ future steps). This is supervised via L2 loss against ground truth actions $a_{t:t+K-1}$ from the demonstration (using last action for padding when near trajectory end). The historical context embedded in $h_t$ enables temporally coherent, long-horizon action generation.}
    \label{fig:mtil_architecture}
\end{figure*}

\subsection{Background and Motivation}
Standard imitation learning often assumes a Markov Decision Process (MDP), learning reactive policies $\pi(a_t | o_t)$ via Behavioral Cloning. However, observational ambiguity fundamentally renders many sequential tasks as Partially Observable MDPs (POMDPs) \cite{osa2018algorithmic}, where the optimal policy necessitates conditioning on the full history $H_t = (o_1, a_1, ..., o_t)$. Theoretically, this history is captured by the belief state $b_t = P(s_t | H_t)$, dictating the optimal policy $\pi^*(a_t | b_t)$ \cite{kaelbling1998planning}.

{Directly computing or representing the belief state $b_t$ is generally intractable. This motivates learning a compressed history representation $h_t \approx b_t$ using recurrent models. This aligns with the core ideas of both World Models, which learn a predictive latent state of the world, and Dynamical Systems (DS) approaches to control, which rely on an evolving internal state. State Space Models (SSMs) like Mamba \cite{gu2023mamba} offer a particularly compelling synthesis of these ideas. They provide a structured recurrent update $h_t = f(h_{t-1}, x_t)$ (where $x_t$ encodes $o_t$) with linear time complexity $O(L)$. This efficiency is the critical enabler for tractably encoding the long sequences required for full history representation, overcoming the computational barriers of prior architectures \cite{gu2021efficiently}.}

{Our motivation for MTIL stems from leveraging Mamba's state $h_t$ as this potent, efficiently computed representation of the full history. We view $h_t$ as the state of a learned, implicit dynamical system that acts as an action-oriented world model. By conditioning actions on both the current observation $o_t$ and this history-infused state $h_t$, MTIL learns a non-Markovian policy:
$$\hat{a}_{t:t+K-1} \approx \pi(o_t, h_t)$$
thereby directly addressing the core challenge of decision-making under ambiguity in POMDP-structured imitation learning by effectively utilizing the entire history.}

\subsection{Leveraging Full Trajectory History with Mamba-2}
MTIL employs Mamba-2~\cite{dao2024transformers}, an advanced structured State Space Model (SSM) notable for its refined selective mechanism and efficiency~\cite{gu2023mamba}. Improving upon Mamba, Mamba-2 enhances hardware utilization and clarifies theoretical links to attention while retaining dynamic context adaptation via input-dependent parameters~\cite{dao2024transformers}. Its core lies in the discretized SSM recurrence governing the hidden state $h_t \in \mathbb{R}^N$ evolution based on input $x_t$ (derived from observation $o_t$):
$$ h_t = \bar{\mathbf{A}}_t h_{t-1} + \bar{\mathbf{B}}_t x_t $$
$$ y_t = \mathbf{C}_t h_t $$
{Crucially, the input-dependent parameters $(\Delta_t, \bar{\mathbf{B}}_t, \mathbf{C}_t) = f(x_t)$ enable selective state dynamics. This allows the model to learn a highly non-linear dynamical system where the system matrices themselves adapt based on the current input. This selective mechanism, combined with inherent linear-time complexity $O(L)$, facilitates learning from \textit{complete} trajectory histories—a significant advantage over quadratic-complexity $O(L^2)$ attention mechanisms. The resulting state $h_t$ acts as a dynamic summary of the salient history $(x_1, ..., x_{t-1})$, furnishing the requisite context even when the instantaneous observation $x_t$ is ambiguous. The MTIL policy leverages this directly:
$$ \pi(\hat{a}_{t:t+K-1} | x_t, h_t) $$
By conditioning predictions $\hat{a}_{t:t+K-1}$ on both the current input $x_t$ and the comprehensive historical summary encoded in $h_t$, MTIL effectively transcends the limitations inherent in Markovian approaches, enabling sequential decision-making grounded in the full trajectory context.}

\begin{algorithm}[!htbp]
\caption{MTIL Training (Sequential Step-based)}
\label{alg:mtil_training}
\small
\begin{algorithmic}[1]
\REQUIRE Expert trajectories $\mathcal{D} = \{ \tau_i \}$, $\tau_i = (o_1, a_1, ..., o_{T_i}, a_{T_i})$, MTIL Policy $\pi_\theta$, initialized parameters $\theta$, Loss $\mathcal{L}$ (MSE), Chunk size $k$, Optimizer $Opt$
\STATE Initialize policy parameters $\theta$
\FOR{each training epoch}
    \FOR{each trajectory $\tau_i \in \mathcal{D}$}
        \STATE Initialize hidden state $h_{0}$, trajectory loss $\mathcal{L}_{traj} \leftarrow 0$
        \FOR{$t = 0$ to $T_i - 1$}
            \STATE $x_t = \text{Encoder}(o_t)$
            \STATE Predict $(\hat{a}_{t:t+k-1}, h_t) = \pi_\theta.\text{step}(x_t, h_{t-1})$
            \STATE Get $a_{t:t+k-1}$ from $\tau_i$
            \STATE Calculate step loss: $\mathcal{L}_t = \mathcal{L}(\hat{a}_{t:t+k-1}, a_{t:t+k-1})$
            \STATE Accumulate loss: $\mathcal{L}_{traj} += \mathcal{L}_t$
        \ENDFOR
        \STATE $Opt.\text{step}()$ \COMMENT{Update $\theta$}
    \ENDFOR
\ENDFOR
\RETURN Trained policy $\pi_\theta$.
\end{algorithmic}
\end{algorithm}

\begin{algorithm}[H]
\caption{MTIL Inference (with Action Chunking and Temporal Aggregation)}
\label{alg:mtil_inference}
\small
\begin{algorithmic}[1]
\REQUIRE Trained policy $\pi_\theta$, Initial observation $o_0$, Chunk size $k$, Max steps $T_{max}$, Exponential aggregation weights $W$
\STATE Initialize hidden state $h_{0}$, prediction buffer $B$.
\FOR{$t = 0$ to $T_{max} - 1$}
    \STATE $x_t = \text{Encoder}(o_t)$
    \STATE Predict $(\hat{a}_{t:t+k-1}, h_t) = \pi_\theta.\text{step}(x_t, h_{t-1})$
    \STATE Store prediction $\hat{a}_{t:t+k-1}$ in buffer $B$
    \STATE Aggregate Action for step $t$:
    \STATE \quad Get predictions for step $t$ from $B$: $P_t = \{ \hat{a}_{j:j+k-1}[t-j] \mid j \le t < j+k \text{ and } \hat{a}_{j:j+k-1} \in B \}$
    \STATE \quad Compute final action: $a_t^{\text{final}} = \text{WeightedAverage}(P_t, W)$
    \STATE Execute action $a_t^{\text{final}}$
\ENDFOR
\end{algorithmic}
\end{algorithm}

\subsection{MTIL Training and Inference}
\label{subsec:training_inference}
{MTIL enables imitation learning across complete expert trajectories, utilizing the architecture outlined in Figure~\ref{fig:mtil_architecture}. Distinctively, MTIL employs a sequential training procedure (Algorithm~\ref{alg:mtil_training}). This step-wise paradigm, leveraging Mamba's recurrent `step` function, is fundamental to efficiently encoding arbitrarily long trajectories from high-dimensional observations (e.g., images) within feasible memory constraints—a key departure from parallel window-based approaches.A naive implementation of this sequential process would be limited to a batch size of one, posing a challenge for training efficiency. To address this, we introduce a novel \textbf{batch-parallel training scheme}. Instead of processing a single trajectory, our method processes a batch of $B$ trajectories simultaneously. At each timestep $t$, the model takes a batch of observations, updates their respective hidden states and computes the loss concurrently. This approach preserves the crucial temporal integrity within each trajectory while fully leveraging the parallel processing power of modern GPUs, making MTIL's training time competitive with highly-parallelizable Markovian methods.During training, at each timestep $t$, the policy receives the observation embedding $x_t$, updates its history-encoding state from $h_{t-1}$ to $h_t$, and predicts an action chunk $\hat{a}_{t:t+K-1}$. Learning proceeds by minimizing the Mean Squared Error (MSE) against the ground truth actions $a_{t:t+K-1}$. During inference (Algorithm~\ref{alg:mtil_inference}), the trained policy operates autoregressively, using the same `step` function to update its state and predict actions. For enhanced stability, temporal aggregation strategies~\cite{shi2023awe, fu2024chunking, zhao2023act} are applied, averaging over predictions from overlapping action chunks to produce a smoother final action.}

\section{Experimental Results}
\label{sec:experiments}
{We conducted extensive experiments to evaluate the performance of MTIL across various benchmarks and real-world scenarios.All results stem from a rigorous protocol over three random seeds (100, 200, 300), with 50 roll-outs and the checkpoint for each run selected based on the minimum validation loss or as the final success rate for Robomimic.}

\subsection{ACT benchmark}
\begin{table}[!htbp]
    \centering
    \captionsetup{type=table, font=footnotesize, skip=3pt}
    {\caption{Success Rates (\%) on the ACT Benchmark. Results are averaged over 3 seeds, All experiments run on a single RTX 4090.}
    \vspace{1mm} 
    \label{tab:act_comprehensive_final}}
    {
    \resizebox{\columnwidth}{!}{%
    \begin{tabular}{@{}lccc@{}}
        \toprule
        \textbf{Method} & \textbf{History Length} & \textbf{Cube Transfer (\%)} & \textbf{Bimanual Insertion (\%)} \\
        \midrule
        ACT~\cite{zhao2023act} & 1 (Markovian) & 90.0 $\pm$ 2.0 & 50.0 $\pm$ 3.5 \\
        \midrule
        Diffusion Policy~\cite{chi2023diffusionpolicy} & 1 (Markovian) & 72.0 $\pm$ 2.6 & 28.0 $\pm$ 3.2 \\
        Diffusion Policy~\cite{chi2023diffusionpolicy} & 10 & 78.0 $\pm$ 2.5 & 32.0 $\pm$ 4.1 \\
        Diffusion Policy~\cite{chi2023diffusionpolicy} & 20 & 80.0 $\pm$ 2.2 & 34.0 $\pm$ 3.8 \\
        Diffusion Policy~\cite{chi2023diffusionpolicy} & 30 & 82.0 $\pm$ 2.1 & 36.0 $\pm$ 3.5 \\
        Diffusion Policy~\cite{chi2023diffusionpolicy} & 40 & \multicolumn{2}{c}{OOM} \\
        \midrule
        Transformer-XL~\cite{tianci2024transformerxl} & Full (400) & 86.0 $\pm$ 2.5 & 42.0 $\pm$ 4.0 \\
        MTIL (10-step) & 10 & 92.0 $\pm$ 1.5 & 56.0 $\pm$ 2.5 \\
        \textbf{MTIL (Full)} & \textbf{Full (400)} & \textbf{100.0 $\pm$ 0.0} & \textbf{84.0 $\pm$ 2.1} \\
        \bottomrule
    \end{tabular}%
    }
    }
\end{table}

\begin{figure*}[!htbp]
    \centering
    \vspace{2mm}
    \begin{minipage}[t]{0.647\linewidth}
        \centering
        \captionsetup{type=figure, font=footnotesize}
        \includegraphics[width=0.98\linewidth]{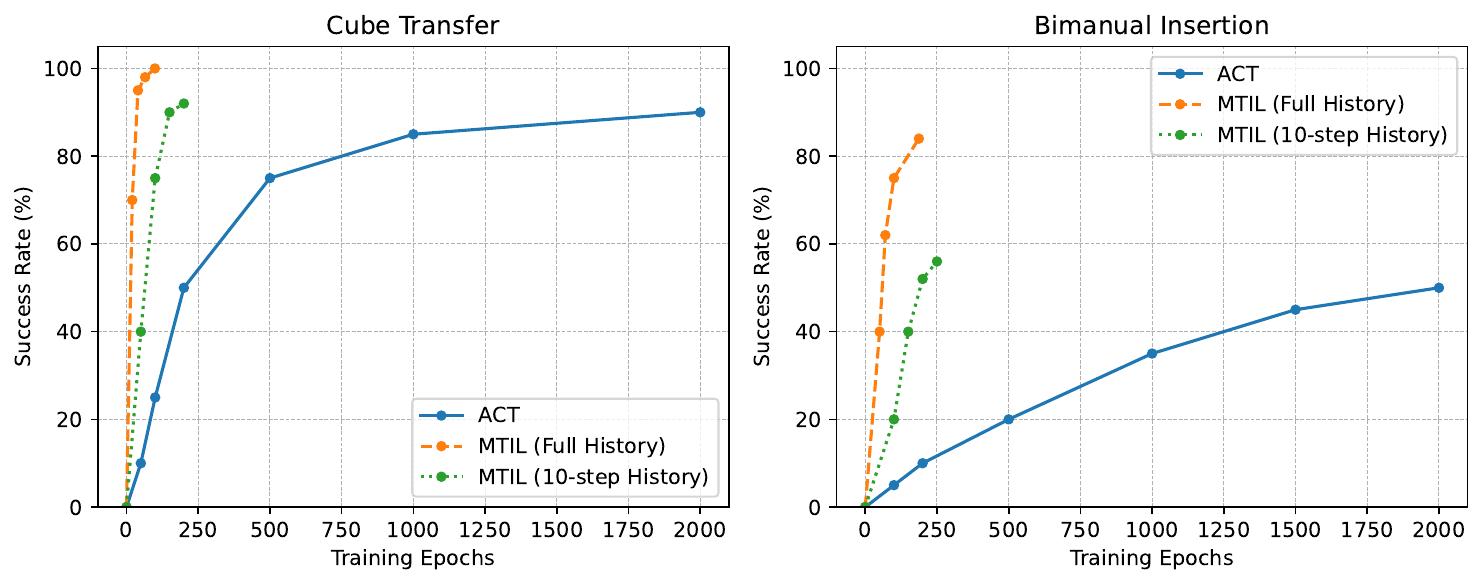}
        \caption*{(a) Learning Curves}
    \end{minipage}%
    \hfill
    \begin{minipage}[t]{0.350\linewidth}
        \centering
        \vspace{1mm}
        \captionsetup{type=figure, font=footnotesize}
        \includegraphics[width=0.95\linewidth]{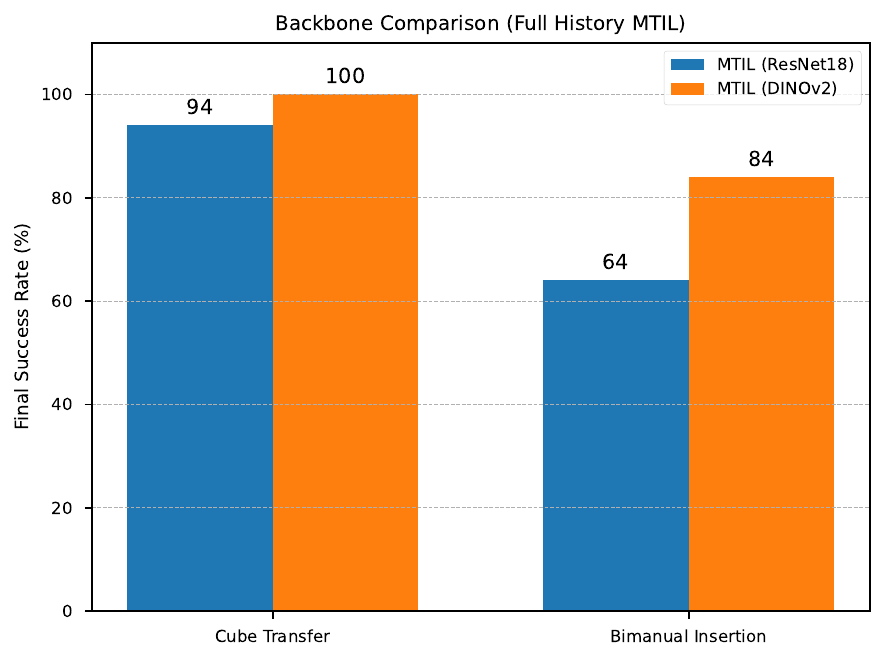}
        \vspace{4mm}
        \caption*{(b) Backbone Comparison}
    \end{minipage}
    \captionsetup{type=figure, font=footnotesize, justification=centering}
    \caption{ACT benchmark performance.(a) Learning curves (Cube Transfer left, Bimanual Insertion right). (b) Backbone comparison (MTIL Full-History, DINOv2 vs. ResNet18).}
    \label{fig:act_plots}
\end{figure*}

{We evaluated MTIL on the ACT benchmark to dissect its performance, efficiency, and learning dynamics against prominent architectural paradigms. The results, which juxtapose success rates with architectural choices and history lengths, are presented in Table~\ref{tab:act_comprehensive_final}.
The findings decisively establish MTIL's superiority. On both tasks, MTIL (Full) achieves a perfect or near-perfect success rate, drastically outperforming all baselines. The learning curves in Figure~\ref{fig:act_plots}(a) illuminate this outcome, showing that MTIL not only attains a higher performance ceiling but also converges significantly faster, indicating a more stable and sample-efficient learning process.
Conversely, the performance of attention-based models reveals a critical insight: naively incorporating history is an inefficient, and ultimately, a computationally infeasible strategy. While Diffusion Policy's success rate scales with history length, it remains notably inferior to the simple Markovian ACT baseline and incurs a prohibitive computational cost, culminating in an Out-of-Memory (OOM) error. Even Transformer-XL, theoretically capable of full-history processing, fails to match ACT, reinforcing the hypothesis that attention is a suboptimal inductive bias for modeling the continuous dynamics of physical interaction.
Furthermore, the backbone ablation in Figure~\ref{fig:act_plots}(b) confirms our advantage is architectural. MTIL, even with an identical ResNet18 backbone~\cite{he2016resnet}, substantially outperforms ACT. The use of a stronger DINOv2 backbone~\cite{oquab2023dinov2} further widens this gap.This proves MTIL's success stems from a fundamentally superior paradigm: a computationally efficient recurrent architecture that is intrinsically better suited to capturing the temporal fabric of the physical world.}

\subsection{LIBERO Benchmark}
On LIBERO's~\cite{liu2023libero} EWC~\cite{kirkpatrick2017ewc} lifelong learning benchmark (using standard ResNet/ViT backbones matching baselines for fair comparison), MTIL demonstrates strong lifelong learning when leveraging full history (\textsc{-M (Full)}, Table~\ref{tab:libero_results}). It consistently achieves superior forward transfer (FWT$\uparrow$), reduced forgetting (NBT$\downarrow$), and higher overall performance (AUC$\uparrow$) compared to baselines and short-history (10-step) MTIL, which performs similarly to Transformers (\textsc{-T}). This advantage of full-history encoding, while notable across all categories, becomes particularly pronounced in \textbf{LIBERO-LONG}. Here, the performance margin over limited-context methods widens substantially , offering compelling evidence for the critical role of complete history as task horizons extend.

\begin{table*}[!htbp]
    \centering
    \captionsetup{type=table, font=scriptsize, skip=3pt}
    \caption{Lifelong Learning Performance on LIBERO (EWC Strategy).}
    \label{tab:libero_results}
    \scriptsize
    \setlength{\tabcolsep}{4pt}
    \begin{tabular}{@{}l ccc @{\hspace{1em}} ccc@{}}
        \toprule
        & \multicolumn{3}{c}{\textbf{LIBERO-LONG}} & \multicolumn{3}{c}{\textbf{LIBERO-SPATIAL}} \\
        \cmidrule(lr){2-4} \cmidrule(lr){5-7}
        Policy Arch. & FWT($\uparrow$) & NBT($\downarrow$) & AUC($\uparrow$) & FWT($\uparrow$) & NBT($\downarrow$) & AUC($\uparrow$) \\
        \midrule
        \textsc{ResNet-RNN} & 0.02 $\pm$ 0.00 & 0.04 $\pm$ 0.01 & 0.00 $\pm$ 0.00 & 0.14 $\pm$ 0.02 & 0.23 $\pm$ 0.02 & 0.03 $\pm$ 0.00 \\
        \textsc{ResNet-T}   & 0.13 $\pm$ 0.02 & 0.22 $\pm$ 0.03 & 0.03 $\pm$ 0.00 & 0.23 $\pm$ 0.01 & 0.33 $\pm$ 0.01 & 0.06 $\pm$ 0.01 \\
        \textsc{ResNet-M (10-step)} & 0.14 $\pm$ 0.02 & 0.20 $\pm$ 0.03 & 0.03 $\pm$ 0.00 & 0.24 $\pm$ 0.01 & 0.30 $\pm$ 0.02 & 0.06 $\pm$ 0.01 \\
        \textsc{ResNet-M (Full)} &\textbf{0.22 $\pm$ 0.03} & 0.08 $\pm$ 0.02 & 0.08 $\pm$ 0.02 & 0.28 $\pm$ 0.02 & 0.17 $\pm$ 0.02 & 0.05 $\pm$ 0.01 \\
        \midrule
        \textsc{ViT-T}      & 0.05 $\pm$ 0.02 & 0.09 $\pm$ 0.03 & 0.01 $\pm$ 0.00 & 0.32 $\pm$ 0.01 & 0.48 $\pm$ 0.03 & 0.06 $\pm$ 0.01 \\
        \textsc{ViT-M (10-step)}    & 0.06 $\pm$ 0.02 & 0.10 $\pm$ 0.03 & 0.01 $\pm$ 0.00 & 0.33 $\pm$ 0.01 & 0.45 $\pm$ 0.03 & 0.06 $\pm$ 0.01 \\
        \textsc{ViT-M (Full)}       & 0.19 $\pm$ 0.04 & \textbf{0.05 $\pm$ 0.01} & \textbf{0.10 $\pm$ 0.03} & \textbf{0.35 $\pm$ 0.02} & \textbf{0.15 $\pm$ 0.03} & \textbf{0.10 $\pm$ 0.01} \\
        \midrule
        \midrule
         & \multicolumn{3}{c}{\textbf{LIBERO-OBJECT}} & \multicolumn{3}{c}{\textbf{LIBERO-GOAL}} \\
        \cmidrule(lr){2-4} \cmidrule(lr){5-7}
        Policy Arch. & FWT($\uparrow$) & NBT($\downarrow$) & AUC($\uparrow$) & FWT($\uparrow$) & NBT($\downarrow$) & AUC($\uparrow$) \\
         \midrule
        \textsc{ResNet-RNN} & 0.17 $\pm$ 0.04 & 0.23 $\pm$ 0.04 & 0.06 $\pm$ 0.01 & 0.16 $\pm$ 0.01 & 0.22 $\pm$ 0.01 & 0.06 $\pm$ 0.01 \\
        \textsc{ResNet-T}   & 0.56 $\pm$ 0.03 & 0.69 $\pm$ 0.02 & 0.16 $\pm$ 0.02 & 0.32 $\pm$ 0.04 & 0.45 $\pm$ 0.04 & 0.07 $\pm$ 0.01 \\
        \textsc{ResNet-M (10-step)} & 0.50 $\pm$ 0.03 & 0.39 $\pm$ 0.03 & 0.15 $\pm$ 0.02 & 0.31 $\pm$ 0.04 & 0.42 $\pm$ 0.04 & 0.07 $\pm$ 0.01 \\
        \textsc{ResNet-M (Full)} & 0.55 $\pm$ 0.03 & 0.36 $\pm$ 0.03 & 0.17 $\pm$ 0.01 & 0.30 $\pm$ 0.03 & 0.11 $\pm$ 0.04 & 0.10 $\pm$ 0.01 \\
        \midrule
        \textsc{ViT-T}      & 0.57 $\pm$ 0.03 & 0.64 $\pm$ 0.03 & 0.23 $\pm$ 0.00 & 0.32 $\pm$ 0.04 & 0.48 $\pm$ 0.03 & 0.07 $\pm$ 0.01 \\
        \textsc{ViT-M (10-step)}    & 0.56 $\pm$ 0.03 & 0.60 $\pm$ 0.03 & 0.22 $\pm$ 0.01 & 0.33 $\pm$ 0.04 & 0.45 $\pm$ 0.03 & 0.08 $\pm$ 0.01 \\
        \textsc{ViT-M (Full)}       & \textbf{0.58 $\pm$ 0.03} & \textbf{0.18 $\pm$ 0.04} & \textbf{0.25 $\pm$ 0.01} & \textbf{0.34 $\pm$ 0.04} & \textbf{0.10 $\pm$ 0.03} & \textbf{0.11 $\pm$ 0.01} \\
        \bottomrule
    \end{tabular}
    \par
    \vspace{2mm}
    \footnotesize FWT($\uparrow$): Forward Transfer, NBT($\downarrow$): Negative Backward Transfer (should be Backward Transfer, if it's negative it's good), AUC($\uparrow$): Area Under Curve. EWC strategy results averaged over 3 seeds (100, 200, 300) at 50 epochs. Baselines from~\cite{liu2023libero}. Short-history (10-step,similar performance for 20/50 steps) and full-history result shown.
\end{table*}

\subsection{Robomimic (Vision-based Policy)}
\begin{table}[H]
    \centering
    \captionsetup{type=table, font=footnotesize}
    {\caption{Behavior Cloning Benchmark (Visual Policy) on Robomimic. As per the original dataset, results are reported as final success rates.\label{tab:robomimic_vision_results_final}}}
    \scriptsize
    \setlength{\tabcolsep}{3pt}
    {
    \begin{tabular}{@{}l|cc|cc|cc|cc|c@{}}
    \toprule
     & \multicolumn{2}{c|}{Lift} & \multicolumn{2}{c|}{Can} & \multicolumn{2}{c|}{Square} & \multicolumn{2}{c|}{Transport} & ToolHang \\
     & ph & mh & ph & mh & ph & mh & ph & mh & ph \\ \midrule
    LSTM-GMM [29] & 1.00 & 1.00 & 1.00 & 0.98 & 0.82 & 0.64 & 0.88 & 0.44 & 0.68 \\
    IBC [12] & 0.94 & 0.39 & 0.08 & 0.00 & 0.03 & 0.00 & 0.00 & 0.00 & 0.00 \\
    \midrule
    DiffusionPolicy-C & \textbf{1.00} & \textbf{1.00} & \textbf{1.00} & \textbf{1.00} & 0.98 & \textbf{0.98} & \textbf{1.00} & 0.89 & 0.95 \\
    DiffusionPolicy-T & \textbf{1.00} & \textbf{1.00} & \textbf{1.00} & \textbf{1.00} & 1.00 & 0.94 & 0.98 & 0.73 & 0.76 \\
    \midrule
    MTIL (10-step) & 1.00 & 1.00 & 1.00 & 0.99 & 0.87 & 0.65 & 0.92 & 0.52 & 0.72 \\
    \textbf{MTIL (Full)} & \textbf{1.00} & \textbf{1.00} & \textbf{1.00} & \textbf{1.00} & \textbf{1.00} & 0.96 & \textbf{1.00} & \textbf{0.91} & \textbf{0.97} \\ \bottomrule
    \end{tabular}
    }
\end{table}
{To assess MTIL's ability to handle high-dimensional visual inputs, we evaluated it on the vision-based Robomimic tasks~\cite{mandlekar2021robomimic}. As shown in Table~\ref{tab:robomimic_vision_results_final}, MTIL (Full) significantly outperforms all baselines, including the strong DiffusionPolicy variants. Notably, MTIL (10-step) offers only a marginal improvement over the LSTM-GMM baseline, highlighting that a short-history SSM is insufficient. The substantial performance gain of MTIL (Full) underscores its superior capability in leveraging full spatio-temporal context from visual data. This can be attributed to its nature as a learned dynamical system; the recurrent state $h_t$ acts as an implicit world model, tracking not just object locations but also their latent physical states (e.g., momentum, contact stability) over time, which is crucial for complex manipulation.}

\subsection{Real-World Dual-Arm Tasks}

\begin{figure}[!htbp]
    \centering
    \captionsetup{type=figure, font=footnotesize}
    \includegraphics[width=0.9\columnwidth]{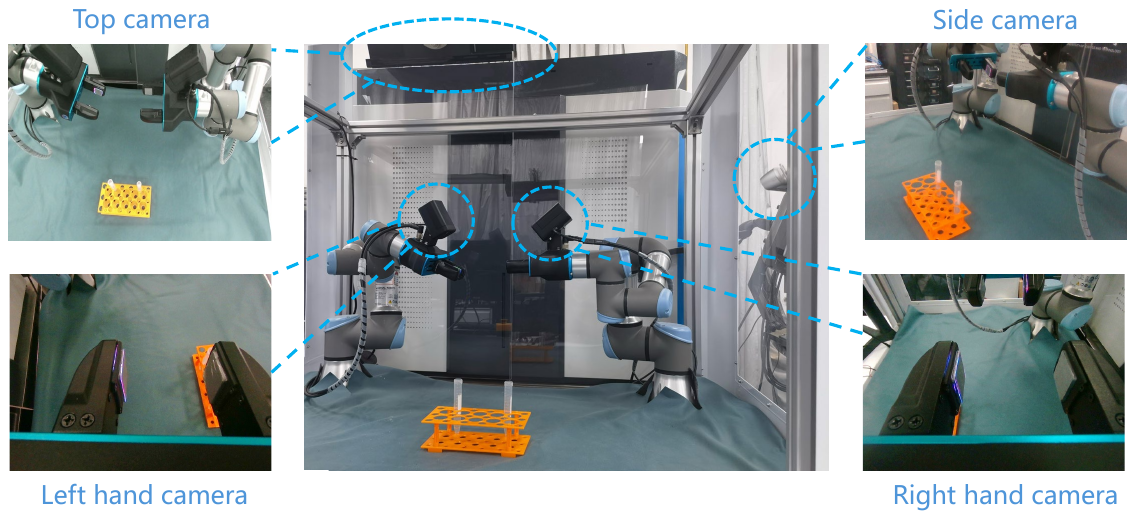}
    \caption{Dual UR3 experimental setup with four cameras (Top: Kinect; Side: D435i; Wrists: D405) and custom grippers.}
    \label{fig:real_setup}
\end{figure}

\begin{figure*}[!htbp]
    \centering
    \vspace{2mm}
    \begin{minipage}[c]{0.6\linewidth}
        \centering
        \includegraphics[width=\linewidth]{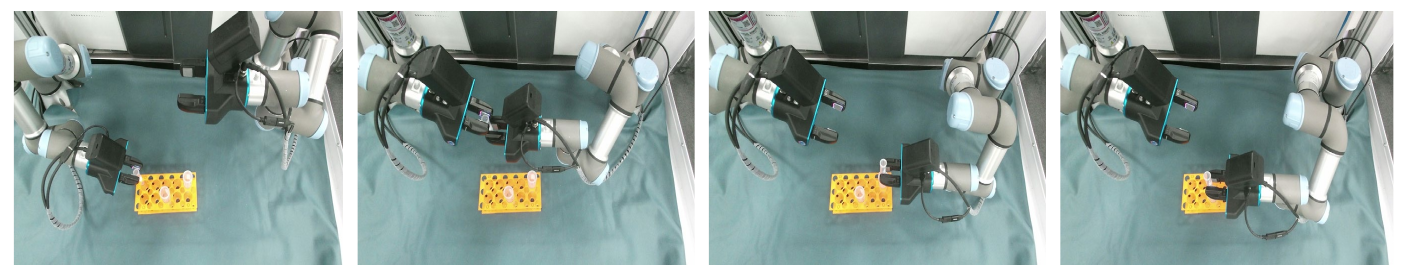}
    \end{minipage}%
    \hfill %
    \begin{minipage}[c]{0.38\linewidth}
        \centering
        \captionsetup{font=footnotesize}
        {\captionof{table}{Sequential Insertion Success Rates (\%), averaged over 50 roll-outs.}
        \label{tab:real_insertion}}
        \footnotesize
        \setlength{\tabcolsep}{2pt}
        {
        \begin{tabular}{@{}lccccc@{}}
            \toprule
            Method & Stage 1 & Stage 2 & Stage 3 & Stage 4 & Overall \\
            \midrule
            ACT    & 80.0$\pm$4.5 & 64.0$\pm$5.1 & 0.0$\pm$0.0  & 0.0$\pm$0.0  & 0.0$\pm$0.0 \\
            MTIL   & \textbf{94.0$\pm$2.5} & \textbf{80.0$\pm$4.0} & \textbf{62.0$\pm$5.8} & \textbf{54.0$\pm$4.5} & \textbf{54.0$\pm$4.5} \\
            \bottomrule
        \end{tabular}
        }
    \end{minipage}
    \captionsetup{font=footnotesize}
    \caption{Sequential Insertion task stages (left panel of figure) and success rates (right panel, Table~\ref{tab:real_insertion}). MTIL successfully completes the sequence while ACT fails due to temporal ambiguity between stages 3 and 4.}
    \label{fig:real_insertion}
\end{figure*}

\begin{figure*}[!htbp]
    \centering
    \begin{minipage}[c]{0.6\linewidth}
        \centering
        \includegraphics[width=\linewidth]{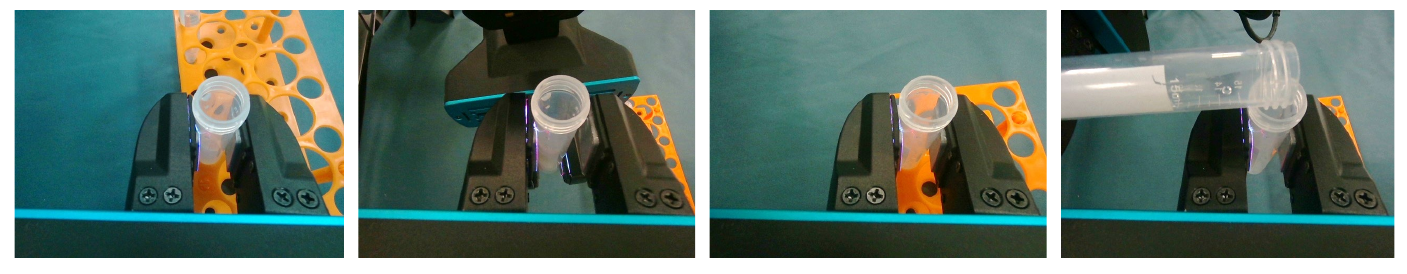}
    \end{minipage}%
    \hfill
    \begin{minipage}[c]{0.38\linewidth}
        \centering
        \captionsetup{font=footnotesize}
        {\captionof{table}{Coordinated Pouring Success Rates (\%), averaged over 50 roll-outs.}
        \label{tab:real_pouring}}
        \footnotesize
        \setlength{\tabcolsep}{2pt}
        {
        \begin{tabular}{@{}lccccc@{}}
            \toprule
             Method & Stage 1 & Stage 2 & Stage 3 & Stage 4 & Overall \\
            \midrule
            ACT    & 80.0$\pm$4.5 & 64.0$\pm$5.1 & 50.0$\pm$6.3 & 32.0$\pm$5.0 & 32.0$\pm$5.0 \\
            MTIL   & \textbf{94.0$\pm$2.5} & \textbf{80.0$\pm$4.0} & \textbf{74.0$\pm$4.8} & \textbf{62.0$\pm$5.8} & \textbf{62.0$\pm$5.8} \\
            \bottomrule
        \end{tabular}
        }
    \end{minipage}
    \captionsetup{font=footnotesize}
    \caption{Coordinated Pouring task stages (left panel of figure) and success rates (right panel, Table~\ref{tab:real_pouring}). MTIL achieves higher success and smoother execution compared to ACT.}
    \label{fig:real_pouring}
\end{figure*}
To validate MTIL in complex physical environments, we designed challenging tasks on a dual UR3 platform equipped with custom 2-finger grippers and four cameras providing multi-view observations (Figure~\ref{fig:real_setup}). We compare MTIL (using DINOv2 backbone and full history) against ACT trained on identical demonstration data (100 demos per task). All real-world results are averaged over 50 evaluation roll-outs for the best checkpoint from each of the 3 seeds.

\paragraph{Sequential Insertion Task.}
We designed this task (visualized in Figure~\ref{fig:real_insertion}) specifically to challenge Markovian policies by requiring long-term memory, a scenario where SOTA methods like ACT often fail. The four stages involve: (1) Left arm grasps Tube1, (2) Left passes Tube1 to Right arm, (3) Right arm inserts Tube1 into Tube2, (4) Right arm inserts Tube1 into Tube3. Critically, executing Stage 3 correctly necessitates recalling the completion of previous stages, as intermediate observations can be ambiguous. Table~\ref{tab:real_insertion} details the stage-wise success rates. MTIL, leveraging its full history state, successfully completes the entire sequence with high probability. In stark contrast, ACT, reliant on immediate context, is confounded by the temporal ambiguity, as the observations after completing Stage 2 can be identical with completing Stage 3, making it indistinguishable for policies relying solely on current or short-term history. AS a result, it frequently attempts Stage 4 directly after Stage 2, failing to execute the required sequence correctly and resulting in zero success for completing Stage 3, Stage 4, and the overall task. This outcome underscores the limitations of short-history approaches and validates the imperative of encoding complete history for reliably executing temporally complex manipulation sequences.
\paragraph{Coordinated Pouring Task.}
This task (Figure~\ref{fig:real_pouring}) assesses precise bimanual coordination over a longer sequence: (1) Left arm grasps Tube1, (2) Left passes Tube1 to Right arm, (3) Left arm grasps Tube2, (4) Right arm pours water from Tube1 into Tube2. While less susceptible to the specific ambiguity of the insertion task, it still requires accurate, temporally coordinated actions. Table~\ref{tab:real_pouring} (within Figure~\ref{fig:real_pouring}) shows that although both methods achieve non-zero success, MTIL consistently outperforms ACT across the stages, resulting in a higher overall success rate and exhibiting notably smoother execution trajectories.

\section{Conclusion}
\label{sec:conclusion}
{The trajectory of intelligence is intrinsically linked to the capacity for memory – the ability to weave the tapestry of past experiences into the fabric of present action. This work confronts a central limitation in contemporary imitation learning: the prevalent reliance on the Markovian assumption, which often reduces complex sequential behaviors to mere reactions to the immediate sensory world. We introduced Mamba Temporal Imitation Learning (MTIL), a new paradigm that embraces the power of memory by leveraging the recurrent state dynamics inherent within the Mamba architecture. We posit that MTIL represents a practical and powerful synthesis of concepts from World Models and Dynamical Systems. By encoding the full history of observations into a compressed, evolving state representation, MTIL learns an implicit, action-oriented world model. This comprehensive temporal context allows MTIL to effectively disambiguate perception and unlock the execution of intricate, state-dependent sequential tasks previously challenging for established methods. Our findings not only showcase the significant performance and efficiency gains afforded by MTIL but, more profoundly, underscore the essential role of history in bridging the gap between perception and intelligent action. By demonstrating the efficacy of SSMs in capturing the long flow of time in a computationally feasible manner, this work illuminates a promising pathway towards building robotic agents capable of deeper understanding and more sophisticated interaction with the world.}

\section*{Acknowledgments}
This work was supported by the Joint Funds of the National Natural Science Foundation of China (Grant No. U222A20208), the Natural Science Foundation Innovation Group Project of Hubei Province (Grant No. 2022CFA018), and the Key Research and Development Program of Guangdong Province (Grant No. 2022B0202010001-2).
\FloatBarrier
\bibliographystyle{IEEEtran}
\bibliography{main}

\end{document}